%% file: main.tex
\PassOptionsToPackage{table, dvipsnames}{xcolor}
\documentclass[10pt,twocolumn,letterpaper]{article}

\usepackage[pagenumbers]{cvpr} 


\definecolor{cvprblue}{rgb}{0.21,0.49,0.74}
\usepackage[pagebackref,breaklinks,colorlinks,allcolors=cvprblue]{hyperref}

\usepackage[accsupp]{axessibility}  
\usepackage{multirow}
\usepackage{multicol}
\usepackage{tipa}

\usepackage[table]{xcolor}
\definecolor{color1}{HTML}{00FF00}
\definecolor{color2}{HTML}{33FF00}
\definecolor{color3}{HTML}{66FF00}
\definecolor{color4}{HTML}{99FF00}
\definecolor{color5}{HTML}{CCFF00}
\definecolor{color6}{HTML}{FFCC00}
\definecolor{color7}{HTML}{FF9900}
\definecolor{color8}{HTML}{FF0000}

\newcommand{\myParagraph}[1]{\noindent \textbf{#1}}

\def\paperID{8182} 
\def\confName{CVPR}
\def\confYear{2025}

\title{SLAM3R: Real-Time Dense Scene Reconstruction from Monocular RGB Videos}

\author{
Yuzheng Liu$^{1}$\thanks{Joint first authors: liu\_yuzheng@stu.pku.edu.cn, siyan3d@hku.hk} \quad
Siyan Dong$^{2}$\footnotemark[1] 
 \footnotemark[2]  \quad
Shuzhe Wang$^{3}$ \quad
Yingda Yin$^{1}$ \\
Yanchao Yang$^{2}$\footnotemark[2] \quad
Qingnan Fan$^{4}$ \quad 
Baoquan Chen$^{1}$\thanks{Corresponding authors} \\
$^1${Peking University} \quad
$^2${The University of Hong Kong} \quad
$^3${Aalto University} \quad 
$^4${VIVO} 
}

\begin{document}
\maketitle

\input{sec/0_abstract}

\input{sec/1_intro}
\input{sec/2_related}

\input{sec/3_method}

\input{sec/4_exp}
\input{sec/5_conclusion}

\input{sec/X_suppl}

{
\small
\bibliographystyle{ieeenat_fullname}
\bibliography{main}
}


\end{document}

%% file: sec/0_abstract.tex
\begin{abstract}

In this paper, we introduce \textbf{SLAM3R}, a novel and effective system for real-time, high-quality, dense 3D reconstruction using RGB videos. SLAM3R provides an end-to-end solution by seamlessly integrating local 3D reconstruction and global coordinate registration through feed-forward neural networks. Given an input video, the system first converts it into overlapping clips using a sliding window mechanism. Unlike traditional pose optimization-based methods, SLAM3R directly regresses 3D pointmaps from RGB images in each window and progressively aligns and deforms these local pointmaps to create a globally consistent scene reconstruction - all without explicitly solving any camera parameters. Experiments across datasets consistently show that SLAM3R achieves state-of-the-art reconstruction accuracy and completeness while maintaining real-time performance at 20+ FPS. Code available at: \url{https://github.com/PKU-VCL-3DV/SLAM3R}.

\end{abstract}

%% file: sec/1_intro.tex
\section{Introduction}
\label{sec:intro}

\begin{figure}[!th]
\centering
\includegraphics[width=0.98\linewidth]{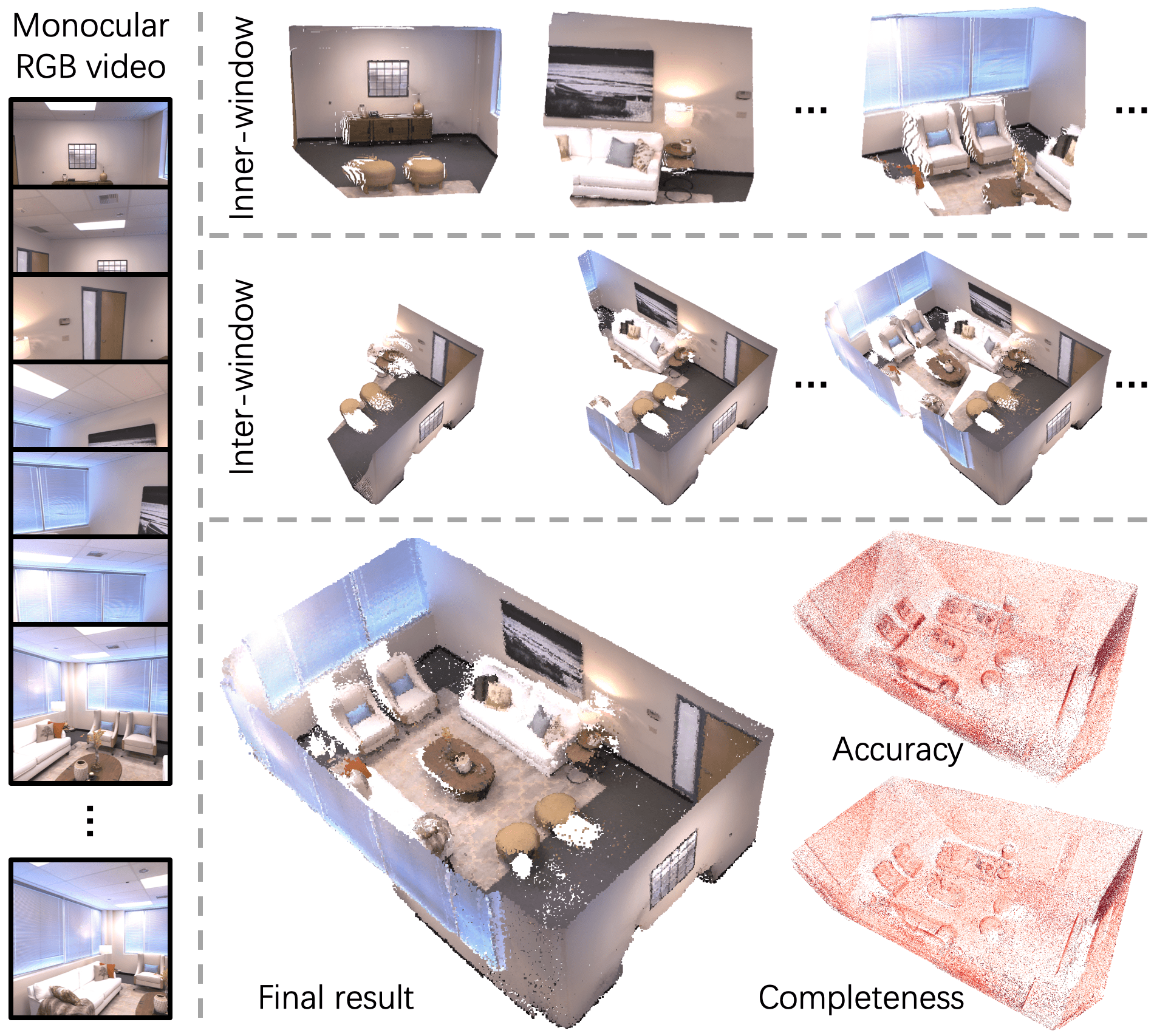}
\caption{
We introduce a novel dense reconstruction system - SLAM3R. 
It takes a monocular RGB video as input and reconstructs the scene as a dense pointcloud. The video is converted into short clips for local reconstruction (denoted as inner-window), which are then incrementally registered together (inter-window) to create a global scene model. This process runs in real-time, producing a reconstruction that is both accurate and complete. 
}
\label{fig:teaser}
\end{figure}

Dense 3D reconstruction, a long-standing challenge in computer vision, aims to capture and reconstruct the detailed geometry of real-world scenes. Traditional approaches have largely relied on multi-stage pipelines. These typically begin with sparse Simultaneous Localization and Mapping (SLAM)~\cite{klein2009parallel,engel2014lsd,mur2015orb,mur2017orb,campos2021orb} or Structure-from-Motion (SfM)~\cite{snavely2006photo,wu2011visualsfm,schonberger2016structure,lindenberger2021pixel,liu20233d} algorithms to estimate camera parameters, followed by Multi-View Stereo (MVS)~\cite{furukawa2015multi,schonberger2016pixelwise,yao2018mvsnet,wang2022itermvs} techniques to fill in scene details. While these methods offer high-quality reconstructions, they often require offline processing to produce a complete model, which limits their applicability in real-world scenarios.

In the literature, dense SLAM approaches~\cite{newcombe2011dtam,zhou2018deeptam,bloesch2018codeslam,czarnowski2020deepfactors,teed2021droid,engel2017direct,sucar2021imap,huang2021di,zhu2022nice,chung2023orbeez,rosinol2023nerf} have been developed to address dense scene reconstruction as a complete system. However, these approaches often fall short in terms of reconstruction accuracy or completeness, or they rely heavily on depth sensors.
Recently, several monocular SLAM systems~\cite{zhang2023go,li2023dense,zhu2024nicer,zhou2024mod,zhang2024glorie,sandstrom2024splat} have been proposed to tackle dense scene reconstruction from RGB videos. By incorporating advanced scene representations~\cite{mildenhall2021nerf,wang2021neus,peng2020convolutional,kerbl20233d,tosi2024nerfs}, these systems produce accurate and complete scene reconstructions. However, this comes at the cost of reduced running efficiency. For example, NICER-SLAM~\cite{zhu2024nicer} operates at a speed significantly below 1 FPS.
Therefore, current approaches struggle with at least one of three key criteria: reconstruction accuracy, completeness, or efficiency.

While monocular dense SLAM systems encounter the limitations mentioned earlier, recent advances in two-view geometry have shown promising potential. DUSt3R~\cite{wang2024dust3r} introduces a purely end-to-end approach for learning dense reconstruction. Trained on large-scale datasets, its network is capable of producing high-quality dense reconstructions from paired images in real-time. However, for multiple views, a global optimization step is required to align these image pairs, which significantly hampers its efficiency. 
A concurrent work, Spann3R~\cite{wang20243d}, extends DUSt3R to multi-view (video) scenarios through an incremental pipeline with spatial memory. While this method accelerates the reconstruction process, it unfortunately results in noticeable accumulated drift and reduced reconstruction quality.

To address these challenges, we present \textbf{SLAM3R} (pronounced ``\textipa{sl\ae m@r}''), 
a real-time dense 3D reconstruction system using RGB-only videos as input. 
Unlike traditional SLAM problems, SLAM3R performs implicit camera localization and focuses more on dense scene mapping, where 3R stands for 3D Reconstruction. 
SLAM3R comprises a two-hierarchy framework. First, it reconstructs local 3D geometry from a sliding window that processes short clips from the input video. Then, it progressively registers these local reconstructions to build a globally consistent 3D scene. Both modules are developed with simple yet effective feed-forward models, enabling end-to-end and efficient scene reconstruction. Specifically, the two modules are the Image-to-Points (I2P) network and the Local-to-World (L2W) network. The I2P module, inspired by DUSt3R, selects a keyframe in a local window as the coordinate system reference. It directly predicts the dense 3D point map supported by the remaining frames within that window. The L2W module incrementally fuses locally reconstructed points into a coherent global coordinate system. 
Both processes reconstruct the 3D points without explicitly estimating any camera parameters. 

Through extensive experiments, we demonstrate that SLAM3R provides high-quality scene reconstructions with minimal drift, outperforming existing dense SLAM systems across various benchmarks. Furthermore, SLAM3R achieves these results at 20+ FPS, bridging the gap between quality and efficiency in RGB-only dense scene reconstruction. Our contributions are summarized below:

\begin{itemize}

\item 
We present a novel real-time end-to-end dense 3D reconstruction system that uses RGB videos to directly predict 3D pointmaps in a unified coordinate system through feed-forward neural networks. 

\item 
Through careful design, our Image-to-Points module can process an arbitrary number of images simultaneously, effectively extending DUSt3R to handle multiple views and produce higher-quality predictions.

\item 
The proposed Local-to-World module directly aligns predicted local 3D pointmaps into a unified global coordinate system. This eliminates the need for explicit camera parameter estimation and costly global optimization.

\item 
We evaluate our method on multiple public benchmarks. It achieves state-of-the-art reconstruction quality in terms of both accuracy and completeness at real-time speeds.

\end{itemize} 

%% file: sec/2_related.tex
\section{Related Work}
\label{sec:related}

\myParagraph{Traditional offline approaches.} 
Dense 3D pointcloud reconstruction is a long-standing problem in computer vision. Classical approaches to this problem first determine camera parameters using Structure from Motion (SfM)~\cite{snavely2006photo,wu2011visualsfm,schonberger2016structure,lindenberger2021pixel,liu20233d}, followed by dense 3D points triangulation with Multi-View Stereo (MVS)~\cite{furukawa2015multi,schonberger2016pixelwise,yao2018mvsnet,wang2022itermvs,aanaes2016large}. 
In recent years, neural implicit~\cite{mildenhall2021nerf,wang2021neus,li2023neuralangelo,wang2023neus2,chen2024sg} and 3D Gaussian~\cite{guedon2024sugar,huang20242d,dai2024high} representations have been applied to further enhance the quality of dense reconstruction. While these methods deliver high-quality results, they have a significant limitation: the requirement for offline processing to generate the final 3D model, which restricts their applicability in real-time scenarios. 
In this paper, we focus on online dense reconstruction in the context of Simultaneous Localization and Mapping (SLAM). 

\myParagraph{Dense SLAM.}
Early works on SLAM~\cite{durrant2006simultaneous,bailey2006simultaneous,klein2009parallel,engel2014lsd,mur2015orb,mur2017orb,campos2021orb} focused on reconstructing the structure of unknown environments while simultaneously localizing camera poses. These approaches prioritize real-time performance but produce only sparse structures of the scene. 
Dense SLAM approaches~\cite{newcombe2011dtam,zhou2018deeptam,bloesch2018codeslam,czarnowski2020deepfactors,koestler2022tandem,teed2021droid,engel2017direct,sucar2021imap,huang2021di,zhu2022nice,chung2023orbeez,rosinol2023nerf,dai2017scannet} incorporate detailed scene geometry information to improve pose estimation. DROID-SLAM~\cite{teed2021droid} introduces recurrent iterative updates of camera poses and pixel-wise depth estimates, while TANDEM~\cite{koestler2022tandem} proposes an online MVS module for depth prediction. These systems enable real-time dense scene reconstruction. However, their focus on camera trajectory accuracy often results in incomplete and noisy 3D reconstruction.
Neural implicit and Gaussian representations have also been integrated with dense SLAM systems~\cite{sucar2021imap,huang2021di,zhu2022nice,rosinol2023nerf,yugay2023gaussian,liso2024loopy,keetha2024splatam,yan2024gs,matsuki2024gaussian,huang2024photo,chung2023orbeez,rosinol2023nerf,sandstrom2023uncle,sandstrom2023point}.  However, these approaches often rely on additional depth sensors or focus primarily on novel view synthesis rather than producing detailed geometric reconstruction. 

More recently, several monocular dense SLAM systems~\cite{zhang2023go,li2023dense,zhu2024nicer,zhou2024mod,zhang2024glorie,sandstrom2024splat} have been developed to produce dense scene geometry reconstruction. A notable limitation of these systems is their slow runtime. Among these systems, GO-SLAM~\cite{zhang2023go} achieves a speed of $\sim$8 FPS, which still falls short of real-time capability.
Furthermore, these methods all share a common strategy: they alternate between solving for camera poses and estimating the scene representation. In contrast, this paper presents a novel approach to dense scene reconstruction that eliminates the need for explicitly solving camera parameters, offering a more efficient and streamlined solution.

\myParagraph{End-to-end dense 3D reconstruction.}
DUSt3R~\cite{wang2024dust3r} introduces the first purely end-to-end dense 3D reconstruction pipeline without relying on camera parameters. Recently, several works have adopted a similar approach for single-view reconstruction~\cite{wang2024moge}, feature matching~\cite{leroy2024grounding}, novel view synthesis~\cite{smart2024splatt3r,ye2024no}, and dynamics reconstruction~\cite{zhang2024monst3r}. These successes demonstrate the effectiveness of end-to-end dense point prediction, inspiring us to develop a dense reconstruction system with a similar methodology.

While DUSt3R operates in real-time for two-view predictions, its extension to multiple views involves exhaustive pairing images and performing an additional global optimization step. This process significantly increases computational time, thereby hindering its real-time performance. 
MASt3R~\cite{leroy2024grounding} enhances the matching capability of DUSt3R by adding a match head, achieving more accurate keypoint correspondences for 3D reconstruction~\cite{duisterhof2024mast3r}, but at the cost of increased computational time. 
More recently, the concurrent work Spann3R~\cite{wang20243d} extends DUSt3R with spatial memory. It takes a video as input and performs incremental scene reconstruction in a unified coordinate system without requiring global optimization. 
While this approach significantly improves runtime efficiency, the incremental reconstruction pipeline frame by frame leads to noticeable accumulated drift.
Unlike Spann3R, our networks at each hierarchy take multiple frames as input to minimize drift. Additionally, we propose a self-contained retrieval module that, when registering a new frame, this module selects not only its previous few frames but also other similar frames from long-term history for more global scene reference.

%% file: sec/3_method.tex
\section{Method}
\label{sec:method}

\begin{figure*}[!th]
\centering
\includegraphics[width=0.99\linewidth]{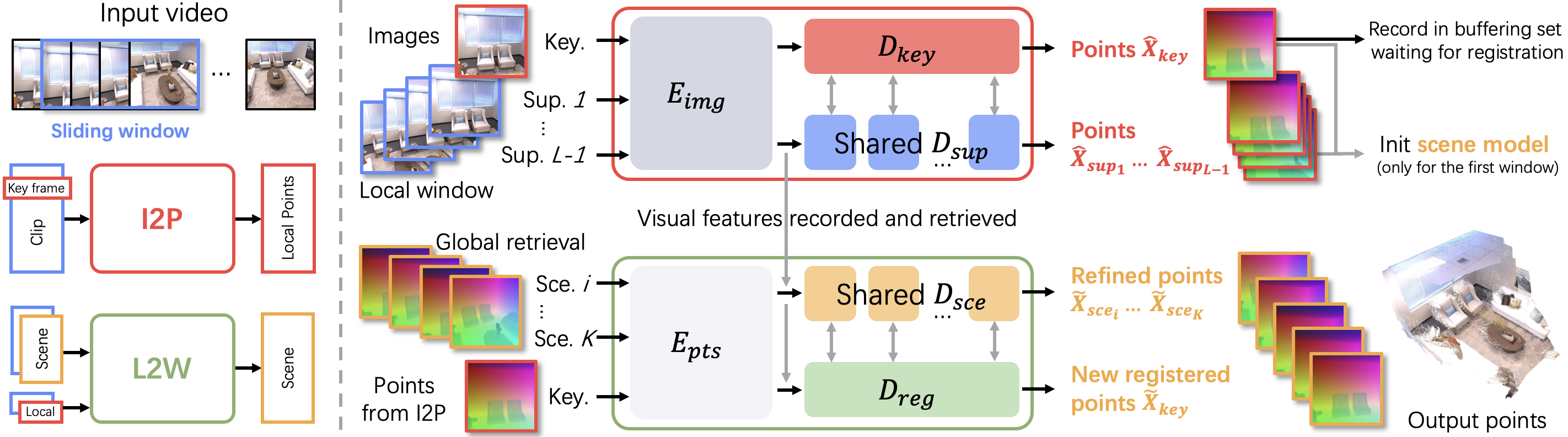}
\caption{
System overview. 
Given an input monocular RGB video, we apply a sliding window mechanism to convert it into overlapping clips (referred to as windows). Each window is fed into an Image-to-Points (I2P) network to recover 3D points in a local coordinate system. Next, the local points are incrementally fed into a Local-to-World (L2W) network to create a globally consistent scene model. The proposed I2P and L2W networks elegantly share similar architectures. 
In the I2P step (Sec.~\ref{sec:inner}), we select a keyframe as a reference to set up a local coordinate system and use the remaining frames in the window to estimate the 3D geometry captured within it. The points from the first window are used to establish the world coordinate system.
We then incrementally fuse the following windows in the L2W step (Sec.~\ref{sec:inter}). This process involves retrieving the most relevant already-registered keyframes as a reference, and integrating new keyframes. Through this iterative process, we eventually obtain the full scene reconstruction. 
}
\label{fig:overview}
\end{figure*}

\myParagraph{Problem statement.}
Given a monocular video consisting of a sequence of RGB image frames $\{I_{i} \in \mathbb{R}^{H\times W \times 3}\}_{i=1}^N$ that captures a static scene, the goal is to reconstruct its dense 3D poincloud $P \in \mathbb{R}^{M \times 3}$, where $M$ is the number of 3D points. 
Our work focuses on three key objectives: maximizing 3D points recovery for reconstruction completeness, improving the accuracy of each recovered point, and achieving these goals while preserving real-time performance.

\myParagraph{System overview.}
Figure~\ref{fig:overview} illustrates an overview of the proposed dense reconstruction system. It consists of two main components: an Image-to-Points (I2P) network that recovers local 3D points from video clips, and a Local-to-World (L2W) network that registers local reconstructions into a global scene coordinate system. 
During the reconstruction of the dense point cloud, the system does not explicitly solve any camera parameters. Instead, it directly predicts 3D point maps in unified coordinate systems.

The system starts by applying a sliding window mechanism of length $L$ to convert the input video into short clips $\{\mathcal{W}_{i} \in \mathbb{R}^{L \times H\times W \times 3}\}$. The I2P network then processes each window $\mathcal{W}_i$ to recover local 3D pointmaps. Within each window, the system selects a keyframe to define a reference coordinate system for point reconstruction, as detailed in Sec.~\ref{sec:inner}. By default, the stride of the sliding window is set to 1, ensuring each input frame in the video is selected at least once as a keyframe. 
For global scene reconstruction, we initialize the world coordinate system with the first window and use the reconstructed frames (image and local point map produced by the I2P) as input for the L2W model. The L2W model incrementally register these local reconstructions into a unified global 3D coordinate system. To ensure both accuracy and efficiency during this process, the system maintains a limited reservoir of registered frames, called scene frames. Whenever the L2W model registers a new keyframe, we retrieve the best-correlated scene frames as a reference. The details are introduced in Sec.~\ref{sec:inter}.

\subsection{Inner-Window Local Reconstruction}
\label{sec:inner}

The Image-to-Points (I2P) model aims to infer dense 3D pointmaps for every pixel of a keyframe in a given video clip. By default, the middle image of a window $\mathcal{W}$ is chosen as the keyframe $I_{key}$ to define the local coordinate system, as it is most likely to have the largest overlap with other frames. The remaining images $\{I_{sup_i}\}_{i=1}^{L-1}$ serve as supporting frames. Note that the 3D pointmaps of supporting frames can also be reconstructed through I2P. 

The I2P network draws inspiration from DUSt3R~\cite{wang2024dust3r}, originally designed for stereo 3D reconstruction. We introduce several simple yet effective modifications to extend it for multi-view scenarios. The I2P model uses a multi-branch Vision Transformer (ViT)~\cite{dosovitskiy2020image} as its backbone. It consists of a shared encoder $E_{img}$, two separate decoders $D_{key}$ and $D_{sup}$, and a point regression head for final prediction. These components are detailed below.

\myParagraph{Image encoder.} 
For a given video clip, the image encoder $E_{img}$ encodes each frame $I_i$ to obtain token representations $F_i \in \mathbb{R}^{T\times d}$, where $T$ is the number of tokens and $d$ is the token dimension. The encoder $E_{img}$ comprises $m$ ViT encoder blocks, each containing self-attention and feed-forward layers. The encoding process is denoted as
\begin{equation*}
F_i^{(T \times d)} = E_{img}(I_i^{(H \times W \times 3)}), \ i=1, ..., L. 
\end{equation*}
The frames are processed independently and in parallel, with the output divided into two parts: $F_{key}$ for the keyframe and $\{F_{sup_i}\}_{i=1}^{L-1}$ for the supporting frames.

\begin{figure}[!tbh]
\centering
\includegraphics[width=0.94\linewidth]{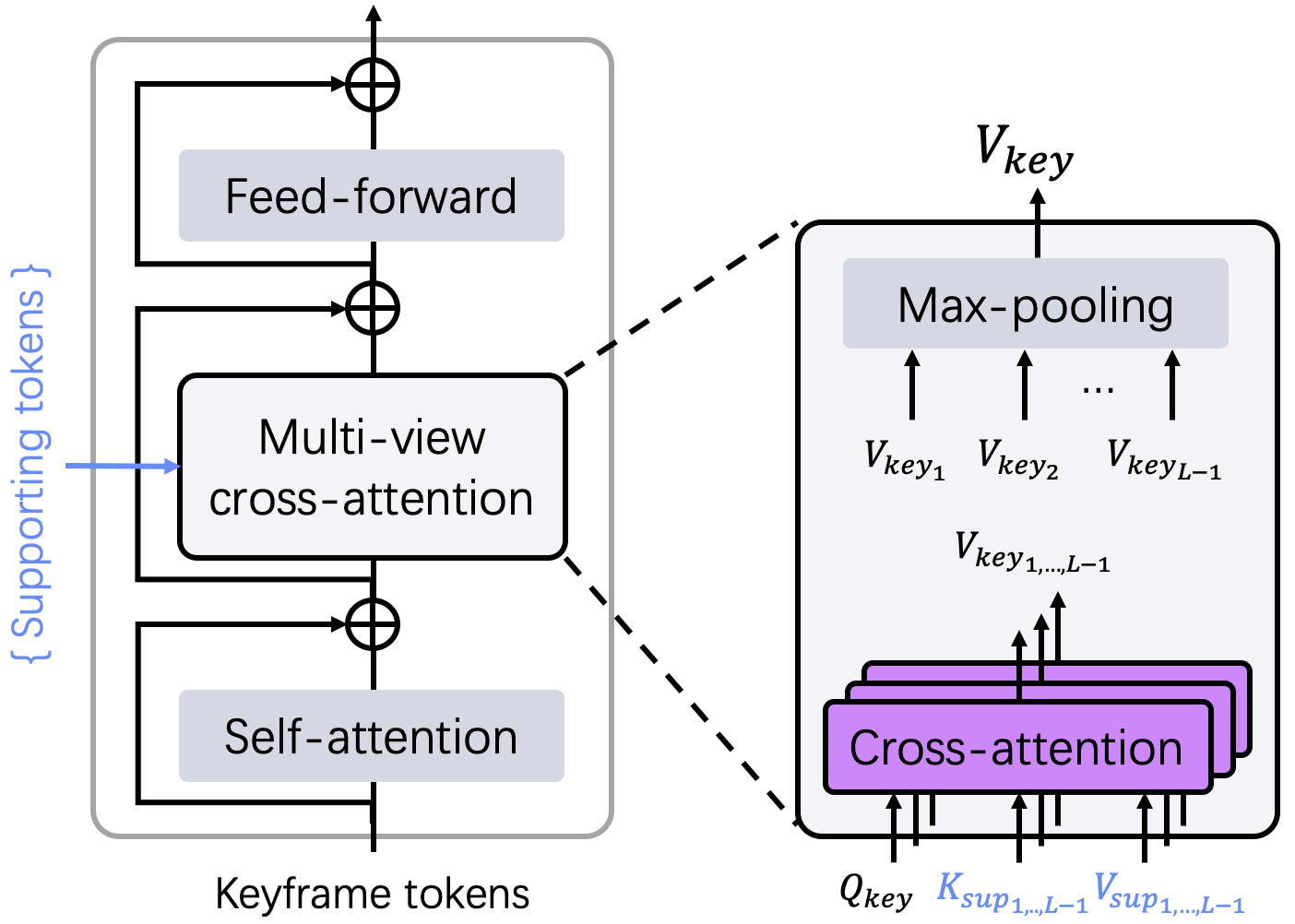}
\caption{
Illustration of a decoder block in the proposed keyframe decoder $D_{key}$. 
We present a minimalist modification to integrate information from different supporting images. Our approach traverses each of them, selects its token keys and values, and uses the keyframe queries to interact with them separately across the supporting images. This multi-view information is then aggregated through max-pooling. 
The registration decoder $D_{reg}$ and scene decoder $D_{sce}$ (described in Sec.~\ref{sec:inter}) share the same architecture.
}
\label{fig:cross-atten}
\end{figure}

\myParagraph{Keyframe decoder.}
The keyframe decoder $D_{key}$ consists of $n$ ViT decoder blocks, each containing self-attention, cross-attention, and feed-forward layers. Unlike DUSt3R which uses the standard cross-attention, we introduce a novel multi-view cross-attention to combine information from different supporting frames. Given the feature tokens $F_{key}$ and $\{F_{sup_i}\}_{i=1}^{L-1}$, the keyframe decoder $D_{key}$ takes $F_{key}$ as input for self-attention and performs cross-attention between $F_{key}$ and $\{F_{sup_i}\}_{i=1}^{L-1}$. A decoder block is illustrated in Figure~\ref{fig:cross-atten}. For each cross-attention layer,  queries are taken from $F_{key}$, while keys and values are extracted from the supporting tokens $F_{sup_i}$. These $L-1$ cross-attention layers are independent of each other, allowing for parallel processing. A max-pooling layer is then employed to aggregate features after cross-attention.  We obtain decoded keyframe tokens $G_{key}$ as:
\begin{equation*}
G_{key} = D_{key}(F_{key}, F_{sup_1}, ..., F_{sup_{L-1}}) .
\end{equation*}

\myParagraph{Supporting decoder.} 
The supporting decoder $D_{sup}$ is designed to complement the keyframe decoder. It inherits the decoder architecture used in DUSt3R, consisting of $n$ standard ViT decoder blocks. The cross-attention mechanism is applied only to exchange information with the keyframe. Note that all supporting frames share the same  $D_{sup}$.
This process is denoted as 
\begin{equation*}
G_{sup_i} = D_{sup}(F_{sup_i}, F_{key}), \ i=1, ..., L-1.
\end{equation*}

\myParagraph{Points reconstruction.} 
Similar to DUSt3R, we apply a linear head~\cite{wang2024dust3r} to regress dense 3D pointmaps in the unified coordinate system from decoded tokens. 
In addition to the pointmaps, we also predict the confidence maps for all frames to evaluate their reliability. The final predictions are:
\begin{equation*}
\hat{X}_i^{(H \times W \times 3)}, \hat{C}_i^{(H \times W \times 1)} = \text{H}(G_i^{(T \times d)}), \ i=1, ..., L. 
\end{equation*}

\myParagraph{Training loss.}
Following DUSt3R, the I2P network is trained end-to-end using ground-truth scene points $\{X_i\}_{i=1}^L$. Both the ground truth and predicted point maps are normalized to a canonical scale, determined by the average distance of all valid points within the window to the origin. The confidence-aware training loss is:

\begin{equation*}
\mathcal{L}_{I2P} = \sum_{i=1}^L M_i \cdot ( \hat{C}_i \cdot \text{L1}(\frac{1}{\hat{z}}\hat{X}_i, \frac{1}{z}X_i) - \alpha \  \text{log} \hat{C}_i ) , 
\end{equation*}
where $M_i$ represents a mask of valid points that have ground-truth values in $X_i$, $z$ and $\hat{z}$ are the scale factor, $\hat{C}_i$ is the confidence map, 
the operator $\cdot$ denotes the element-wise matrix multiplication,
L1$(\cdot)$ denotes the point-wise Euclidean distance, and $\alpha$ is a hyper-parameter to control the regularization term. 
We will detail the process in Sec~\ref{sec:exp}.

\subsection{Inter-Window Global Registration}
\label{sec:inter}

After obtaining the 3D pointmap $\{\hat{X}_{key}\}$ from the I2P network, we use the inter-window Local-to-World (L2W) model to incrementally register the newly generated pointmap into a global 3D coordinate system. 
Similar to the I2P network, the L2W model also relies on some frames to serve as a reference for the scene coordinate system. Furthermore, it can leverage multiple registered keyframes as a global reference.
These registered keyframes are referred to as scene frames and they are maintained in a buffering set through a sampling mechanism.

The buffering set is designed for scalability in handling long videos. We apply a reservoir strategy~\cite{vitter1985random} that maintains a maximum of $B$ registered frames in the buffering set. When a new keyframe is inferred from I2P and ready for fusion, we retrieve the top-$K$ best-correlated scene frames from the buffering set as its support for global registration.

\myParagraph{Scene initialization.} 
The first window is used to initialize the scene model. It's crucial to ensure this initialization is as accurate as possible. 
To achieve this, we execute the I2P network $L$ times, attempting to traverse and designate each frame within the window as the keyframe.
We then select the result with the highest total confidence score for scene model initialization. This process results in a scene pointcloud along with a set of registered frames. All these frames are regarded as scene frames, and they are used to initialize the buffering set.

\myParagraph{Reservoir and retrieval.} 
Each scene frame $I_{sce_i}$ is recorded with its latent feature $F_{sce_i}$ and pointmaps $\tilde{X}_{sce_i}$. For efficiency, we apply reservoir sampling to allow storing an unbiased subset of an empirical distribution in a bounded amount of memory. The first $B$ registered frames chosen are directly inserted into the buffering set. For each subsequent frame with $id>B$, the probability of inserting it is $B/id$. If chosen for insertion, it will randomly replace one of the current scene frames in the buffering set. 

Given a new keyframe $I_{key}$  to be registered, we feed its feature $F_{key}$ and the features from the buffering set into a retrieval module, 
\begin{equation*}
\text{Retrieval}(F_{key}^{(T \times d)}, \{F_{sce_i}^{(T \times d)}\}), 
\end{equation*}
to obtain a list of correlation scores, measuring both the visual similarity and baseline suitability between the keyframe and the scene frames in the buffering set. 
The retrieval module uses the first $r$ decoder blocks from the I2P module as its backbone. A linear projection and an average-pooling layer follow, together producing an image-wise correlation score. We then select the top-$K$ scene frames as a global reference to fuse the current keyframe. As a result, we have $K$ scene frames and one keyframe as the input for the following L2W model.

\myParagraph{Points embedding.} 
The 3D pointmaps reconstructed by the I2P model are encoded into the L2W model using a patch embedding method similar to image patchification in the ViT encoder $E_{img}$. We process the new keyframe and $K$ retrieved scene frames in parallel as:
\begin{equation*}
\mathcal{P}_i^{(T \times d)} = E_{pts}(\hat{X}_i^{(H \times W \times 3)}), \ i=1, ..., K+1. 
\end{equation*}
The encoded geometric tokens are combined with their corresponding visual tokens by
\begin{equation*}
\mathcal{F}_i^{(T \times d)} = F_i^{(T \times d)} +\mathcal{P}_i^{(T \times d)}, \ i=1, ..., K+1. 
\end{equation*}
This resulting a token set $\{\mathcal{F}_{key}, \{\mathcal{F}_{sce_i}\}_{i=1}^K\}$ contains joint features of image patch appearance and 3D geometry for the keyframe and retrieved scene frames. 
In the following decoders, $\{\mathcal{P}\}$ are further accumulated to $\{\mathcal{F}\}$ between adjacent blocks to enhance the geometric representation. 

\myParagraph{Registration decoder.} 
The registration decoder $D_{reg}$ takes feature tokens $\{\mathcal{F}_{key}, \{\mathcal{F}_{sce_i}\}_{i=1}^K\}$ as input and aims to transform the local reconstruction of the keyframe to the scene coordinate system. It takes the same network architecture of the keyframe decoder $D_{key}$. This decoding process is denoted by 
\begin{equation*}
\mathcal{G}_{key} = D_{reg}(\mathcal{F}_{key}, \mathcal{F}_{sce_1}, ..., \mathcal{F}_{sce_{K}}) .
\end{equation*}

\myParagraph{Scene decoder.} 
The scene decoder $D_{sce}$ takes the token set $\{\mathcal{F}_{key}, \{\mathcal{F}_{sce_i}\}_{i=1}^K\}$ as input to refine the scene geometry without coordinate system changes. 
It uses the same network architecture as the keyframe decoder $D_{key}$, allowing us to extend to multi-keyframe co-registration (see supplementary material for details). 
By default, we register one keyframe each time. 
Each of the $\mathcal{F}_{sce_i}$ has information exchange only with the $\mathcal{F}_{key}$. This decoding process is denoted by 
\begin{equation*}
\mathcal{G}_{sce_i} = D_{sce}(\mathcal{F}_{sce_i}, \mathcal{F}_{key}), \ i=1, ..., K.
\end{equation*}

\myParagraph{Points reconstruction and training loss.} 
We apply the same head design as that of the I2P network to predict all the pointmaps $\tilde{X}_i$ in the global scene coordinate system:
\begin{equation*}
\tilde{X}_i^{(H \times W \times 3)}, \tilde{C}_i^{(H \times W \times 1)} = \text{H}(\mathcal{G}_i^{(T \times d)}), \ i=1, ..., K+1. 
\end{equation*}
We train the L2W network using a similar loss function as the I2P network. Differently, no normalization is applied to the predicted point map, as the output scale must align with the scene frames in the input. This alignment ensures that the output can be directly integrated into the existing reconstruction. The training loss of the L2W network is:
\begin{equation*}
\mathcal{L}_{L2W} = \sum_{i=1}^L M_i \cdot ( \tilde{C}_i \cdot \text{L1}(\tilde{X}_i, X_i) - \alpha \  \text{log} \tilde{C}_i ) .
\end{equation*}
The following section provides a detailed discussion of the training process and its implementation details.

%% file: sec/4_exp.tex
\section{Experiments}
\label{sec:exp}

\input{tab/7s}

\input{tab/replica}

\input{tab/pose}

\input{fig/fig_visual_compare}

\input{fig/fig_visual_realworld}

\myParagraph{Datasets.}
For both the Image-to-Points (I2P) and Local-to-World (L2W) models, we perform training with a mixture of three datasets: ScanNet++~\cite{yeshwanth2023scannet++}, Aria Synthetic Environments~\cite{avetisyan2024scenescript} and CO3D-v2~\cite{reizenstein2021common}. These datasets vary from scene-level to object-centric, and contain both real-world and synthetic scenes. Since they are all recorded sequentially, we can easily extract video clips with a sliding-window mechanism as our training data. We select about 850K clips for training in total.
To validate our reconstruction quality, we conduct quantitative evaluations on two unseen datasets: 7 Scenes~\cite{shotton2013scene}, a real-world dataset of partial scenes, and Replica~\cite{replica19arxiv}, a synthetic dataset of complete scenes.   
We also demonstrate visual reconstruction results across diverse datasets and in-the-wild captured videos to showcase the generalization ability of SLAM3R. 

\myParagraph{Implementation details.}
Both the I2P and L2W models build upon the architecture of DUSt3R~\cite{wang2024dust3r} with minimal but effective modifications, making it natural for them to initialize their weights from the DUSt3R pre-trained model. 
We initialize our weights using the DUSt3R model trained on 224$\times$224 resolution images with $m=24$ encoder blocks, $n=12$ decoder blocks with linear heads. 
All images are center-cropped to 224$\times$224 pixels before feeding into SLAM3R. 
Our training is conducted on 8 NVIDIA 4090D GPUs, each with 24 GB of memory. It takes about one day. 
At test time, we set the initial window length to $L=5$ to ensure high-quality reconstruction of all frames within the window. For subsequent incremental windows, we use $L=11$ to provide more supporting views for better keyframe reconstruction.
Please refer to our supplementary material for more implementation details.

\subsection{Comparisons}
\label{sec:comp}

\myParagraph{Evaluation metrics.}
Following NICER-SLAM~\cite{zhu2024nicer} and Spann3R~\cite{wang20243d}, we build a ground-truth point cloud model for each test sequence by back-projecting pixels to the world using ground-truth depths and camera parameters. The reconstructed point clouds are aligned to ground truths using Umeyama~\cite{umeyama1991least} and ICP~\cite{rusinkiewicz2001efficient} algorithms. We quantify reconstruction quality through accuracy and completeness metrics. To demonstrate computational efficiency, we report frames per second (FPS) on a single NVIDIA 4090D GPU. We also evaluate camera poses using absolute trajectory error (ATE-RMSE). For detailed formulations of the metrics, please refer to the supplementary material.

\myParagraph{Reconstruction results on the 7 Scenes~\cite{shotton2013scene} dataset.}
The numerical results of scene reconstruction quality are reported in Table~\ref{tab:7s}. Following Spann3R~\cite{wang20243d}'s setting, we uniformly sample one-twentieth of the frames in each test sequence as input video. Each video is regarded as an individual scene. 
We evaluate SLAM3R using two settings: integrating the full pointmaps predicted for all input frames to create reconstruction results (denoted by SLAM3R-NoConf), and filtering pointmaps with a confidence threshold of 3 before creating reconstruction results (SLAM3R). 
We compare our method with optimization-based reconstruction  DUSt3R~\cite{wang2024dust3r}, triangulation-based MASt3R~\cite{leroy2024grounding}, and online incremental reconstruction Spann3R. 
Our method outperforms all baselines in both accuracy and completeness while maintaining real-time performance. As shown in the Office-09 scene (the top row in Figure~\ref{fig:compare}), our approach demonstrates much less drift compared to the concurrent work Spann3R~\cite{wang20243d}.

\myParagraph{Reconstruction results on the Replica~\cite{replica19arxiv} dataset.} Besides the baselines mentioned in 7 Scene datasets, we also compare the SLAM-based reconstruction approaches NICER-SLAM~\cite{zhu2024nicer}, DROID-SLAM~\cite{teed2021droid}, DIM-SLAM~\cite{li2023dense} and GO-SLAM~\cite{zhang2023go} on the Replica~\cite{replica19arxiv} dataset. 
The numerical results on full scene reconstruction are reported in Table~\ref{tab:replica}. 
Due to the memory constraint, DUSt3R~\cite{wang2024dust3r} and MASt3R~\cite{leroy2024grounding} process only one-twentieth of the frames for reconstruction. 
As is shown in the table, our method surpasses all baselines with FPS greater than 1. Notably, without any optimization procedure, our method achieves reconstruction quality comparable to optimization-based methods such as NICER-SLAM~\cite{zhu2024nicer} and DUSt3R~\cite{wang2024dust3r}. Example of the Office 2 (the bottom row in Figure~\ref{fig:compare}) also illustrates the global consistency of our reconstruction result. 

\myParagraph{Camera pose estimation on 7 Scenes~\cite{shotton2013scene} and Replica~\cite{replica19arxiv}.} 
Our method is designed in a new paradigm that reconstructs 3D points end-to-end without explicitly solving camera parameters. Following DUSt3R~\cite{wang2024dust3r}, We also derive camera poses from the predicted scene points using PnP-RANSAC solver in OpenCV~\cite{bradski2000opencv} with ground truth camera intrinsics of each frame.  
The results are reported in Table~\ref{tab:pose}. 
We can observe that camera poses and scene reconstruction results are not fully positively correlated. This discrepancy between pose and reconstruction errors indicates that effective end-to-end 3D reconstruction is possible and promising without first obtaining precise camera poses.

For more details on the comparisons in this section, please refer to our supplementary material.

\subsection{Analyses}
\label{sec:analyses}

\myParagraph{Effectiveness of the I2P model.}
To highlight the advantages of our multi-view I2P model over the original two-view DUSt3R~\cite{wang2024dust3r}, we evaluate the reconstruction quality of keyframes with varying numbers of supporting views. We conduct experiments on the Replica~\cite{replica19arxiv} dataset, where input views are sampled using a sliding window of different sizes, and the reconstruction accuracy and completeness of the keyframes are computed. The results are reported in Table~\ref{tab:local_window}. As the number of supporting views increases, our approach progressively improves reconstruction quality. Notably, the efficiency of our method remains stable until the window size exceeds 11, demonstrating the effectiveness of our parallel design. However, the results also show diminishing returns as the number of views increases, which we detail in the supplementary material. 
Visual results of I2P reconstruction can be found in Figure~\ref{fig:teaser}.

\input{tab/local_window}

\input{tab/align_method}

\myParagraph{Advantages of the L2W model.}
The effectiveness of the L2W model is evaluated through ablation studies on the Replica~\cite{replica19arxiv} dataset. Per-window reconstructions are first generated with a window size of 11 using the I2P model. Local points are then aligned to a unified coordinate frame using different methods: global optimization from DUSt3R~\cite{wang2024dust3r} (I2P-GA), traditional approaches such as Umeyama~\cite{umeyama1991least} and ICP~\cite{rusinkiewicz2001efficient} (I2P+UI), and our L2W model (I2P+L2W+Re). For consistency, we set the window size for global optimization to 10, which is equal to the number of views used to align new frames in other methods. Results in Table~\ref{tab:align_method} show that our full method achieves superior alignment accuracy and computational efficiency compared to the alternatives.

\myParagraph{Analysis of the retrieval module.} 
We propose a lightweight retrieval module that selects historical scene frames from the reservoir. This approach effectively performs implicit re-localization. We compare our retrieval method with a baseline approach that selects the ten nearest previous frames, named I2P+L2W. The results in Table \ref{tab:align_method} indicate a significant performance improvement with our retrieval strategy, demonstrating its effectiveness.

\myParagraph{In-the-wild scene reconstruction.}
We have tested our method on a diverse range of unseen datasets and found that SLAM3R shows strong generalization capabilities. Figure~\ref{fig:in-the-wild} shows our reconstruction results on Tanks and Temples~\cite{knapitsch2017tanks}, BlendedMVS~\cite{yao2020blendedmvs}, Map-free Reloc~\cite{arnold2022map}, LLFF~\cite{mildenhall2019local}, and ETH3D~\cite{schops2019bad,schops2017multi} datasets, as well as in-the-wild videos we captured. 
These results show that our method performs reliably across different scales and scenes.
We also provide additional numerical results on sampled scenes from these datasets in the supplementary material. 

%% file: tab/7s.tex
\begin{table*}[th!]
\centering

\resizebox{0.93\textwidth}{!}{

\small

\begin{tabular}{l|ccccccc|c|c} 
\toprule
 \multirow{2}{*}{Method} & Chess & Fire & Heads & Office & Pumpkin & RedKitchen & Stairs & Average & \multirow{2}{*}{FPS} \\ 

 & Acc. / Comp. & Acc. / Comp.& Acc. / Comp. & Acc. / Comp. & Acc. / Comp. & Acc. / Comp. & Acc. / Comp. & Acc. / Comp. \\ 

\toprule

DUSt3R~\cite{wang2024dust3r} & 2.26 / 2.13 & \textbf{1.04} / 1.50  & 1.66 / \textbf{0.98} & 4.62 / 4.74 & \textbf{1.73 / 2.43} & \textbf{1.95 / 2.36} & 3.37 / \textbf{10.75} & \textbf{2.19 / 3.24} & \cellcolor{color7!25} $<$1 \\ 

MASt3R~\cite{leroy2024grounding} & \textbf{2.08 / 2.12} & 1.54 / \textbf{1.43} & \textbf{1.06} / 1.04 & \textbf{3.23 / 3.19} & 5.68 / 3.07 & 3.50 / 3.37 & \textbf{2.36} / 13.16 & 3.04 / 3.90 & \cellcolor{color8!25} $\ll$1 \\ 

\midrule
    
Spann3R~\cite{wang20243d} & 2.23 / 1.68 & 0.88 / 0.92 & \textbf{2.67 / 0.98} & 5.86 / 3.54 & 2.25 / \textbf{1.85} & 2.68 / \textbf{1.80} & 5.65 / \textbf{5.15} & 3.42 / 2.41 & \cellcolor{color1!25} $>$50 \\

\textbf{SLAM3R-NoConf (Ours)} & 2.12 / \textbf{1.21} & 0.95 / \textbf{0.80} & 3.23 / 1.67 & 2.59 / \textbf{2.21} & 1.99 / 2.04 & 2.09 / 1.88 & 4.54 / 6.38 & 2.40 / \textbf{2.24} & \cellcolor{color2!25} $\sim$25\\

\textbf{SLAM3R (Ours)} & \textbf{1.63} / 1.31 & \textbf{0.84} / 0.83 & 2.95 / 1.22 & \textbf{2.32} / 2.26 & \textbf{1.81} / 2.05 & \textbf{1.84} / 1.94 & \textbf{4.19} / 6.91 & \textbf{2.13} / 2.34 & \cellcolor{color2!25} $\sim$25 \\

\bottomrule
\end{tabular}
}
\caption{
Reconstruction results on 7 Scenes \cite{shotton2013scene} dataset.
{
The average numbers are computed over all test sequences. 
The methods are categorized into two groups based on whether their FPS is above or below 1.
The best results within each category are shown in bold. 
We report accuracy and completeness in centimeters. 
The color gradient shifts from red through yellow to green to show increasing FPS. 
} 
}
\label{tab:7s}
\end{table*}

%% file: tab/replica.tex
\begin{table*}[th!]
\centering
\resizebox{1\textwidth}{!}{

\small 

\begin{tabular}{l|cccccccc|c|c}
\toprule

\multirow{2}{*}{Method}& Room 0 & Room 1 & Room 2 & Office 0 & Office 1 & Office 2 & Office 3 & Office 4 & Average & \multirow{2}{*}{FPS} \\ 

& Acc. / Comp. & Acc. / Comp. & Acc. / Comp. & Acc. / Comp. & Acc. / Comp. & Acc. / Comp. & Acc. / Comp. & Acc. / Comp. & Acc. / Comp. & \\ 

\toprule

{DUSt3R}~\cite{wang2024dust3r} & 3.47 / \textbf{2.50} & \textbf{2.53} / \textbf{1.86} & \textbf{2.95} / \textbf{1.76} & 4.92 / 3.51 & \textbf{3.09} / \textbf{2.21} & 4.01 / 3.10 & \textbf{3.27} / \textbf{2.25} & 3.66 / 2.61 & \textbf{3.49} / \textbf{2.48} & \cellcolor{color7!25} $<$1 \\ 

MASt3R~\cite{leroy2024grounding} & 4.01 / 4.10 & 3.61 / 3.25 & 3.13 / 2.15 & \textbf{2.57} / \textbf{1.63} & 12.85 / 8.13 & \textbf{3.13} / \textbf{1.99} & 4.67 / 3.15 & 3.69 / \textbf{2.47} & 4.71 / 3.36 & \cellcolor{color8!25} $\ll$1 \\

NICER-SLAM~\cite{zhu2024nicer}* & \textbf{2.53} / 3.04 & 3.93 / 4.10 & 3.40 / 3.42 & 5.49 / 6.09 & 3.45 / 4.42 & 4.02 / 4.29 & 3.34 / 4.03 & \textbf{3.03} / 3.87 & 3.65 / 4.16 & \cellcolor{color8!25} $\ll$1 \\ 

\midrule

DROID-SLAM~\cite{teed2021droid}* & 12.18 / 8.96 & 8.35 / 6.07 & 3.26 / 16.01 & \textbf{3.01} / 16.19 & \textbf{2.39} / 16.20 & 5.66 / 15.56 & 4.49 / 9.73 & 4.65 / 9.63 & 5.50 / 12.29 & \cellcolor{color3!25} $\sim$20 \\

DIM-SLAM~\cite{li2023dense}* & 14.19 / 6.24 & 9.56 / 6.45 & 8.41 / 12.17 & 10.16 / 5.95 & 7.86 / 8.33 & 16.50 / 8.28 & 13.01 / 6.77 & 13.08 / 8.62 & 11.60 / 7.85 & \cellcolor{color5!25} $\sim$3 \\

GO-SLAM~\cite{zhang2023go} & - & - & - & - & - & - & - & - & 3.81 / 4.79 & \cellcolor{color4!25} $\sim$8 \\ 

Spann3R~\cite{wang20243d} & 9.75 / 12.94 & 15.51 / 12.94  & 7.28 / 8.50 & 5.46 / 18.75 & 5.24 / 16.64 & 9.33 / 11.80 & 16.00 / 9.03 & 13.97 / 16.02 & 10.32 / 13.33 & \cellcolor{color1!25} $\textgreater$50 \\

\textbf{SLAM3R-NoConf (Ours)} & 3.37 / \textbf{2.40} & 3.22 / \textbf{2.33}  & 3.15 / \textbf{2.00} & 4.43 / \textbf{2.59}  & 3.18 / \textbf{2.34} & 3.95 / \textbf{2.78} & 4.20 / \textbf{3.15}  & 4.57 / 3.38 & 3.76 / \textbf{2.62} & \cellcolor{color2!25} $\sim$24 \\

\textbf{SLAM3R (Ours)} & \textbf{3.19} / \textbf{2.40} & \textbf{3.12} / 2.34 & \textbf{2.72} / \textbf{2.00} & 4.28 / 2.60 & 3.17 / \textbf{2.34} & \textbf{3.84} / \textbf{2.78} & \textbf{3.90} / 3.16 & \textbf{4.32} / \textbf{3.36} & \textbf{3.57} / \textbf{2.62} & \cellcolor{color2!25} $\sim$24 \\

\bottomrule
\end{tabular}

}
\caption{
Reconstruction results on Replica~\cite{replica19arxiv} dataset.
* denotes the results reported in NICER-SLAM.
}
\label{tab:replica}
\end{table*}

%% file: tab/pose.tex
\begin{table*}[htb!]
\centering
\resizebox{0.97\textwidth}{!}{
\begin{tabular}{l|ccc|cccccc} 
\toprule
& 
DUSt3R~\cite{wang2024dust3r} & 
MASt3R~\cite{leroy2024grounding} & 
NICER-SLAM~\cite{zhu2024nicer}* & 
DROID-SLAM~\cite{teed2021droid}* & 
DIM-SLAM~\cite{li2023dense} & 
GO-SLAM~\cite{zhang2023go} & Spann3R~\cite{wang20243d} & 
\textbf{SLAM3R-NoConf (Ours)} & \textbf{SLAM3R (Ours)} \\  

\toprule

7 Scenes &
8.02 & 
\textbf{6.28} & 
8.55 & 
\textbf{5.66} & - & - & 
11.70 & 
8.44 & 8.41 \\

\midrule

Replica & 
4.76 & 
\textbf{1.67} & 
1.88 & 
\textbf{0.33} & 0.46 & 0.39 &
32.79 &
6.61 & 6.61\\
\bottomrule
\end{tabular}
}
\caption{
Camera pose results evaluated by ATE-RMSE (cm) on 7 Scenes~\cite{shotton2013scene} and Replica~\cite{replica19arxiv} datasets. 
}
\label{tab:pose}
\end{table*}

%% file: fig/fig_visual_compare.tex
\begin{figure*}[!th]
\centering
\includegraphics[width=0.99\linewidth]{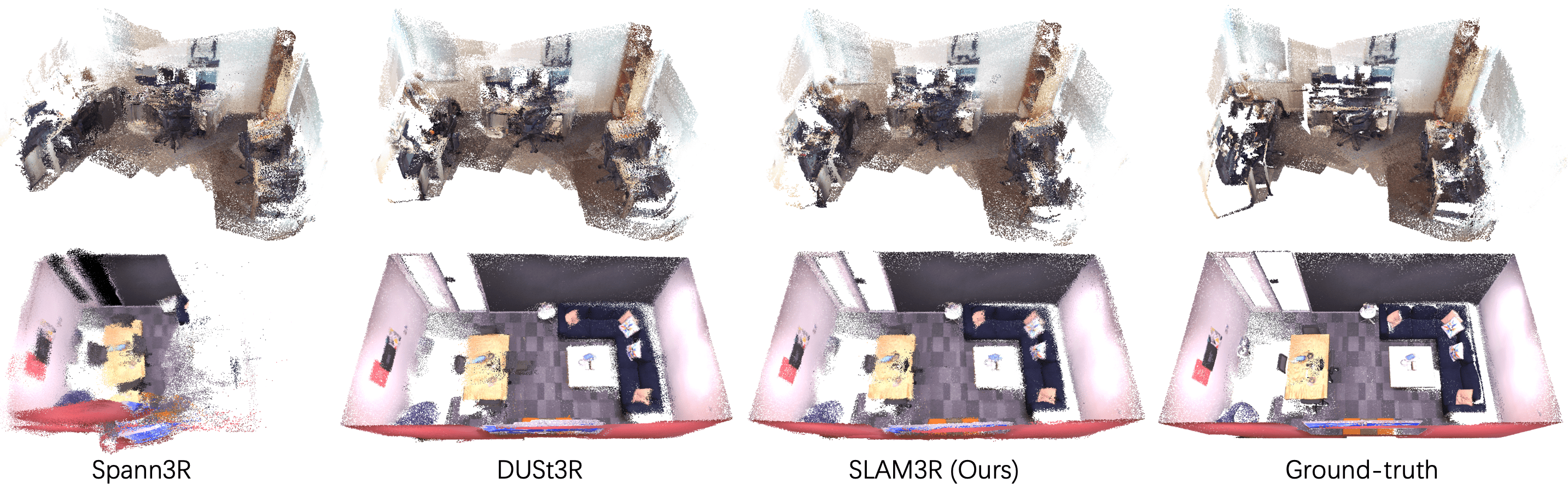}
\caption{
We visualize the reconstruction results on two scenes: Office-09 and Office 2 from the 7-Scenes~\cite{shotton2013scene} and Replica~\cite{replica19arxiv} datasets. Our method runs in real-time and achieves high-quality reconstruction comparable to the offline method DUSt3R~\cite{wang2024dust3r}.
}
\label{fig:compare}
\end{figure*}

%% file: fig/fig_visual_realworld.tex
\begin{figure*}[!th]
\centering
\includegraphics[width=0.99\linewidth]{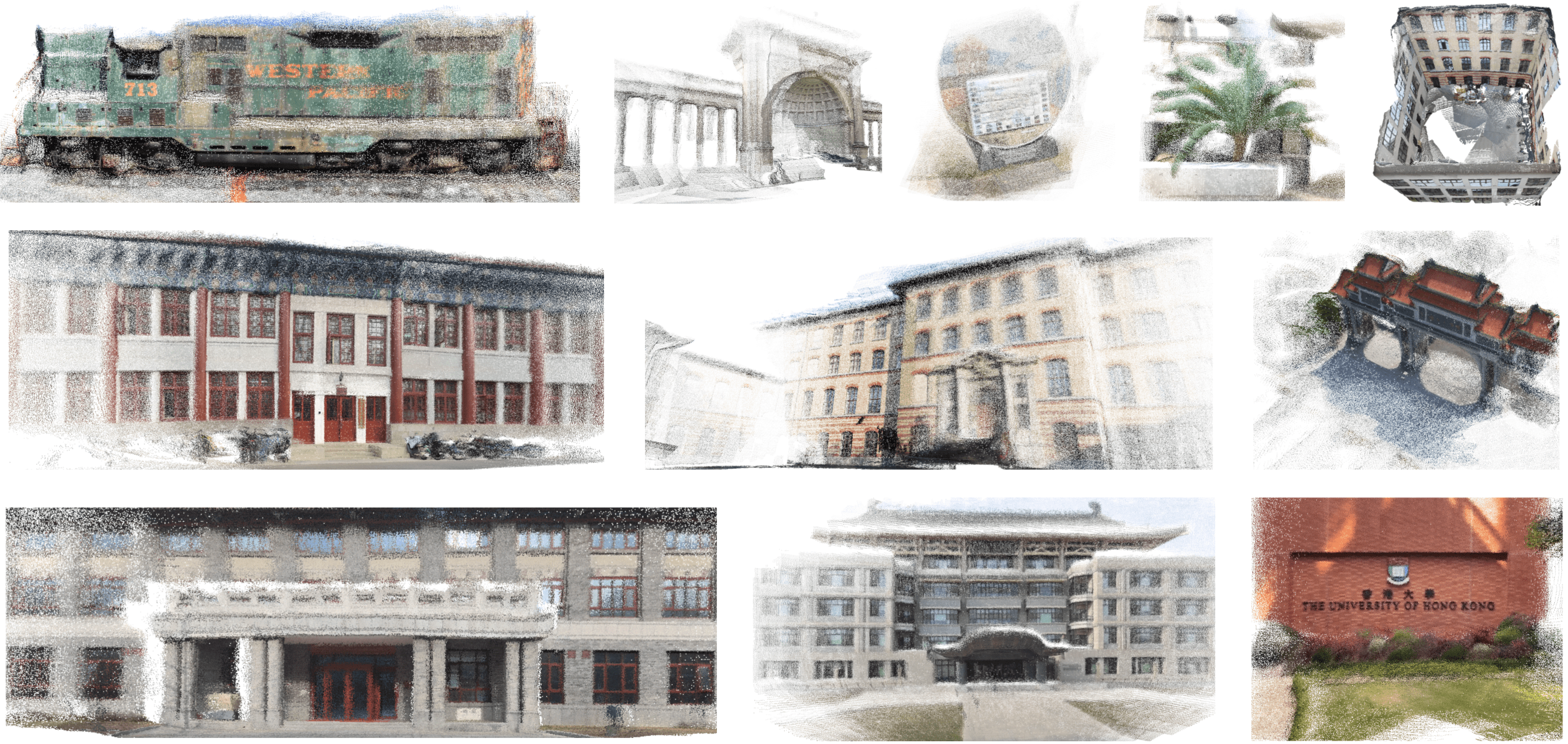}
\caption{
{Qualitative examples.} 
We show our reconstruction results on Tanks and Temples~\cite{knapitsch2017tanks}, BlendedMVS~\cite{yao2020blendedmvs}, Map-free Reloc~\cite{arnold2022map}, LLFF~\cite{mildenhall2019local}, and ETH3D~\cite{schops2019bad,schops2017multi} datasets, as well as in-the-wild captured videos, to demonstrate SLAM3R's generalization ability.
}
\label{fig:in-the-wild}
\end{figure*}

%% file: tab/local_window.tex
\begin{table}
    \centering
    \footnotesize

    \begin{tabular}{lcccc}
    \toprule
        Method & \# Frames & Acc. & Comp. & FPS  \\
        \toprule
        DUSt3R~\cite{wang2024dust3r} & 2 & 3.16 & 2.89 & \cellcolor{color1!25} 42.55 \\
        
        \midrule
        
        {I2P} & 2 & 3.39 & 3.04 & \cellcolor{color1!25} 42.55 \\

        {I2P} & 5 & 2.62 & 2.28 & \cellcolor{color3!25} 40.82 \\

        {I2P (Default)} & 11 & 2.38 & 2.03 & \cellcolor{color4!25} 40.11 \\

        {I2P} & 15 & 2.27 & 1.94 & \cellcolor{color5!25} 35.51 \\
        {I2P} & 51 & \textbf{2.23} & \textbf{1.86} & \cellcolor{color7!25} 11.97 \\
    \bottomrule
    \end{tabular}
\caption{
Inner-window keyframe reconstruction results with various window lengths. 
By default, we use 11-frame windows for incremental reconstruction to balance quality and efficiency.
}
    \label{tab:local_window}
\end{table}

%% file: tab/align_method.tex
\begin{table}
    \centering
    \footnotesize

    \begin{tabular}{lccc}
    \toprule
        Method & Acc. & Comp. & FPS  \\
        \toprule 
    
        I2P+GA & 4.87 & 3.00 & \cellcolor{color5!25} $\sim$3 \\
        
        {I2P+UI} & 7.47 & 3.86 & \cellcolor{color7!25} $\sim$1 \\
        {I2P+L2W} & 6.19 & 3.54 & \cellcolor{color1!25} $\sim$92 \\
        {I2P+L2W+Re (Full)} & \textbf{3.62} & \textbf{2.70} & \cellcolor{color3!25} $\sim$43 \\
    \bottomrule
    \end{tabular}
\caption{
Reconstruction results using various point alignment methods and scene frame selection strategies. The FPS reported only accounts for the overhead of the alignment operation.
}
    \label{tab:align_method}
\end{table}

%% file: sec/5_conclusion.tex
\section{Conclusion}
\label{sec:conclusion}

In this paper, we present SLAM3R, a novel and effective system that performs real-time, high-quality, dense 3D scene reconstruction using RGB videos. 
It employs a two-hierarchy neural network framework to perform end-to-end 3D reconstruction through streamlined feed-forward processes, eliminating the need to explicitly solve any camera parameters. Experiments demonstrate its state-of-the-art reconstruction quality and real-time efficiency (20+ FPS).  

\noindent
\textbf{Limitations and future work.}
The elimination of camera parameter prediction prevents us from performing global bundle adjustment. Additionally, the poses derived from our scene point cloud prediction still fall short of SLAM systems that specialize in camera localization. Addressing these limitations will be the focus of our future work.

\section{Acknowledgments}
This work is supported by the National Key R\&D Program of China (2022ZD0160800).
This work is also supported by 
the Early Career Scheme of the Research Grants Council (grant \# 27207224),
the HKU-100 Award, 
a donation from the Musketeers Foundation, 
and in part by the JC STEM Lab of Robotics for Soft Materials funded by The Hong Kong Jockey Club Charities Trust.
We thank the reviewers for their valuable feedback, as well as Zihan Zhu and Songyou Peng for their help with our experiments. 
Siyan Dong would also like to thank the support from HKU School of Computing and Data Science.

%% file: sec/X_suppl.tex


\appendix
\section*{Supplementary Material}

In this appendix, we first present additional implementation details and experimental settings in Sec.~\ref{sec:more_imp} and Sec.~\ref{sec:more_exp}, which were omitted from the main paper due to page limit. We then report additional analyses in Sec.~\ref{sec:more_ana}. Finally, we show more reconstruction results of our method in Sec.~\ref{sec:more_vis}.

\section{Implementation details}
\label{sec:more_imp}
\paragraph{Retrieval module.}
We propose a lightweight module for efficient scene frame retrieval to support the keyframe registration. 
The retrieval module directly reuses I2P's decoder blocks as its backbone, followed by a linear projection and an average-pooling layer. Specifically, it uses the first two blocks from both the supporting and keyframe decoders, for scene frames and keyframes (awaiting registration), respectively. 
It takes as input image features of one keyframe and all the scene frames in the buffering set, predicting correlation scores between the keyframe and each buffering frame. 
Notably, the correlation scores share similar behavior with the mean confidence of the I2P model's final prediction and offer unique advantages over the cosine similarity between image features of two frames. These correlation scores account for both visual similarity and provide suitable baselines for 3D reconstruction.

The module inherits the weights of the first two layers of the decoder in I2P model. During training, only the weights of the linear projection are updated using an L1 loss:
\begin{align*}
\mathcal{L}_{Retr} &=  \sum_{i=1}^{R}\lvert S_i' - Mean(C_i') \rvert , \\
S_i'&= Sigmoid(S_i) , \\[2mm]
C_i' &=  (C_i -  1)/C_i ,
\end{align*}
where R is the number of input supporting frames, $S_i$ is the predicted correlation score between supporting frame i and the keyframe, $C_i$ is the predicted confidence from the complete I2P model. Both $S_i$ and $C_i$ are normalized to [0,1] before calculating the loss.

\paragraph{Multi-keyframe co-registration.}
In practice, our scene decoder in the L2W model adopts the same architecture as the keyframe decoder in the I2P model, allowing for the simultaneous input and registration of multiple keyframes. In the decoding stage, scene frames and keyframes exchange information bidirectionally: each scene frame queries features from all keyframes, and each keyframe interacts with all scene frames. Compared to single-keyframe registration, this extension significantly reduces computational overhead by registering multiple keyframes with a single pass of the scene decoder. Furthermore, incorporating information from additional keyframes enhances the refinement of scene frame features, leading to more accurate reconstruction for all input frames.

\paragraph{Training details.}
To construct the training data, we utilize all iPhone and DSLR frames registered by COLMAP~\cite{schonberger2016structure} from the training splits of ScanNet++\cite{yeshwanth2023scannet++}. Additionally, we include all frames from the first 450 scenes of the Aria Synthetic Environments (ASE)\cite{avetisyan2024scenescript} dataset and 41 categories from CO3D-v2~\cite{reizenstein2021common}, with each category containing up to 50 randomly sampled scene sequences.
We introduce two ways to extract video clips for training.
For ScanNet++ and ASE, we adopt uniform sampling with strides of 3 and 2, respectively. For CO3D-v2, frames are randomly sampled within temporal segments covering half the length of each video. In total, we extract approximately 850K clips. 
During each epoch of training, we randomly sample 4000, 2000, and 2000 clips from the ScanNet++, ASE, and CO3D-v2 datasets, respectively. All training images are resized and then center-cropped to 224 × 224 pixels. Standard data augmentation techniques~\cite{wang2024dust3r} are applied. 

To train our I2P model, we extend the training process of DUSt3R from two views to multiple views. Specifically, our I2P model takes as input a video clip of length 11, and designates the middle frame as the keyframe. We train the I2P model for 100 epochs, which takes about 6 hours. After that, we train the retrieval module built on the I2P model. During training, we freeze all other modules and use L1 loss to supervise the correlation score against the mean confidence of the I2P model's final predictions. This module requires 50 epochs of training, which takes about 2 hours.

To train the L2W model, we use clips of length 12, with the first six images selected as scene frames, and the last six images designated as keyframes to register. The model is trained for 200 epochs in total, and the training process takes approximately 16 hours. 
When training with ground truth pointmaps as input, we set invalid points to (0,0,0). A confidence-aware loss without scale normalization is applied, ensuring that the predicted point maps retain consistent scale with the input scene frames. 

Our training is conducted on 8 NVIDIA 4090D GPUs, each with 24GB of memory and a batch size of 4 per GPU.

\input{fig/master_failure}
\input{fig/incremental}
\input{fig/dtu}
\input{tab/7s_cam}
\input{tab/replica_cam}

\section{Details for experimental settings}
\label{sec:more_exp}

\paragraph{Calculation of the evaluation metrics.}
To evaluate reconstruction quality, we use accuracy and completeness as our metrics. They are calculated by:
$$ Accuracy= \frac{1}{P}\sum_{i=1}^P min_j(D(x_i, y_j)), $$
$$ Completeness= \frac{1}{Q}\sum_{j=1}^Q min_i(D(x_i, y_j)). $$
$P$ and $Q$ are the numbers of points in the reconstructed point cloud and GT point cloud respectively. $D(\cdot)$ represents Euclidean distance, and $x_i$ and $y_j$ represent iterating each point from the reconstructed and GT point cloud. 

To measure the efficiency, we report FPS (frames per second), which is calculated by: 
$$ FPS = F/time, $$
where $time$ is the total time used to reconstruct the scene, and $F$ is the number of frames from the video. 

We evaluate the camera pose accuracy using absolute trajectory error (ATE-RMSE), which is formulated by: 
$$ ATE\mbox{-}RMSE = \sqrt{\frac{1}{F}\sum_{i=1}^F D(T^{gt}_i, T^{perd}_i)^2}, $$
where $T^{perd}$ and $T^{gt}$ are the camera center positions of the predicted and GT camera trajectories.

\paragraph{Full video as input on Replica~\cite{replica19arxiv}.}
On the Replica dataset, we reconstruct the entire scene geometry using all video frames. With the stride of the sliding window set to 1, all frames will be used as a keyframe once. 
For each window, frames are sampled around the keyframe, with $Skip=20$ frames per supporting frame, to ensure reasonable camera motion (disparity). 
We co-register $Co=10$ keyframes at each time, which share the same $K=10$ scene frames as a reference. These scene frames are selected through a two-step process. First, we calculate the correlation score between all frames in the buffering set and the $Co$ keyframes. Then, we select $K$ frames from the buffering set that show the highest total correlation score with these keyframes.
After every $R=20$ registered keyframes, we update the buffering set by retaining the keyframes with the highest reconstruction scores, where reconstruction score of a frame is the product of its mean confidence predicted by I2P and L2W model. 
The insertion/update follows the reservoir sampling probability described in the main paper.

\paragraph{Sampled frames as input on 7 Scenes~\cite{shotton2013scene}.}
Following Spann3R~\cite{wang20243d}, the frames in each test sequence are sampled with a stride of 20, and we only reconstruct the points from the sampled frames. 
To handle sampled-frame-only input, we adapt our reconstruction pipeline for full-video input by setting $Skip=1$, $Co=2$, $K=5$, and $R=1$ in practice.

\paragraph{Experiments on DUSt3R~\cite{wang2024dust3r} and MASt3R~\cite{leroy2024grounding}.} 
The global optimization with complete graph setting in DUSt3R and MASt3R requires substantial GPU memory. Consequently, to evaluate the global reconstruction quality of these two methods on the Replica dataset, we uniformly sample 1/20 of the images. 
DUSt3R is tested using the weight-224 model with a resolution of 224\(\times\)224, the same as our input resolution, while MASt3R is tested using the weight-512 model with resolutions of 512$\times$384 and 512$\times$288 as inputs for reconstructing the 7 Scenes~\cite{shotton2013scene} and Replica~\cite{replica19arxiv} datasets, respectively. 
Note that a resolution of 224$\times$224 results in less overlap between adjacent frames, making reconstruction inherently more challenging. 

During the evaluation, we observed that MASt3R occasionally generates floating points with high confidence scores, which are difficult to filter using confidence thresholds and significantly degrade accuracy. An example of this issue is shown in Figure~\ref{fig:master_failure}. 
In contrast, our confidence scores are more effective and successfully reduce erroneous points. The results of SLAM3R reported on 7 Scenes and Replica datasets use a fixed confidence threshold of 3.

\section{Additional comparisons and analyses}
\label{sec:more_ana}

\paragraph{More numerical results.} 
We report more quantitative comparisons of reconstruction results on ScanNet~\cite{dai2017scannet}, Tanks and Temples~\cite{knapitsch2017tanks}, and ETH3D~\cite{schops2019bad} datasets. We sampled three scenes from each dataset, and report the results in Table~\ref{tab:reb_scannet}. SLAM3R outperforms Spann3R in most cases and demonstrates performance either comparable to or better than DUSt3R. These results further verify our method’s effectiveness.

\input{tab/scannet}

\paragraph{Diminishing return of window length.}
\input{fig/fig_diminishing_return}
In the main paper, we report the I2P reconstruction results with different window lengths. Here, we further analyze the diminishing returns, which indicate that the window length should not be too large. As Figure~\ref{fig:diminishing_return} shows, the accuracy and completeness of the keyframe reconstruction improve rapidly at first as input frames increase, but then gradually decline. 
This is because larger windows result in less and less overlapping. Additionally, the inference time becomes significantly slower as length increases. 
Consequently, we set the window size to 11 in our main experiments, balancing the reconstruction quality and runtime efficiency. 

\paragraph{Effect of scene frame numbers on registration.}
\input{tab/sceneframe_num}
We conduct experiments on the Replica~\cite{replica19arxiv} dataset to investigate how the number of scene frames selected as a global reference affects the registration quality of keyframes. As reported in Table~\ref{tab:scene_frame_num}, the accuracy of full-scene registration initially improves as the maximum number of input scene frames increases but eventually declines beyond a certain threshold. 
Retrieving too few scene frames from the buffering set risks missing suitable frames and causing keyframe registration to get stuck in local minimums. Conversely, selecting too many scene frames can introduce irrelevant ones that add noise and hinder registration.

To balance reconstruction accuracy and runtime efficiency, we set the number of retrieved scene frames to 5 and 10 on 7 Scenes~\cite{shotton2013scene} and Replica~\cite{replica19arxiv} dataset, which achieves consistent and reliable performance.

\paragraph{Camera pose estimation.}
The detailed results are presented in Table~\ref{tab:7s_cam} and Table~\ref{tab:replica_cam}. 
For DUSt3R~\cite{wang2024dust3r} and MASt3R~\cite{leroy2024grounding}, we evaluate the camera poses derived via the PnP-RANSAC solver with their predicted pointmaps (after global alignment) and GT intrinsic parameters. When evaluating Spann3R~\cite{wang20243d} on the Replica~\cite{replica19arxiv} dataset, only one-twentieth of the frames are used, as it fails to give reasonable results with all frames input. 

We outperform the concurrent work Spann3R~\cite{wang20243d}, demonstrating the effectiveness of our hierarchical design with multi-view input and global retrieval. 
Among classical SLAM systems, the pose errors of GO-SLAM~\cite{zhang2023go} and DROID-SLAM~\cite{teed2021droid} are lower than those of NICER-SLAM. However, their reconstruction accuracy and completeness are worse. 
This discrepancy between pose and reconstruction errors indicates that effective end-to-end 3D reconstruction is possible and promising without first obtaining precise camera poses. 

\section{More visual results}
\label{sec:more_vis}

\paragraph{Visualization of incremental reconstruction.}
Figure~\ref{fig:incremental} visualizes the process of our incremental reconstruction on two scenes from Replica~\cite{replica19arxiv}. Our method achieves effective alignment at loops while experiencing minimal cumulative drift, without offline global optimization step.

\paragraph{Reconstruction on DTU~\cite{aanaes2016large} dataset.}
The results are shown in Figure~\ref{fig:dtu}. Note that our method does not require any camera parameters, and produces dense point cloud reconstructions end-to-end in real-time.

%% file: fig/master_failure.tex
\begin{figure*}[!th]
\centering
\includegraphics[width=0.98\linewidth]{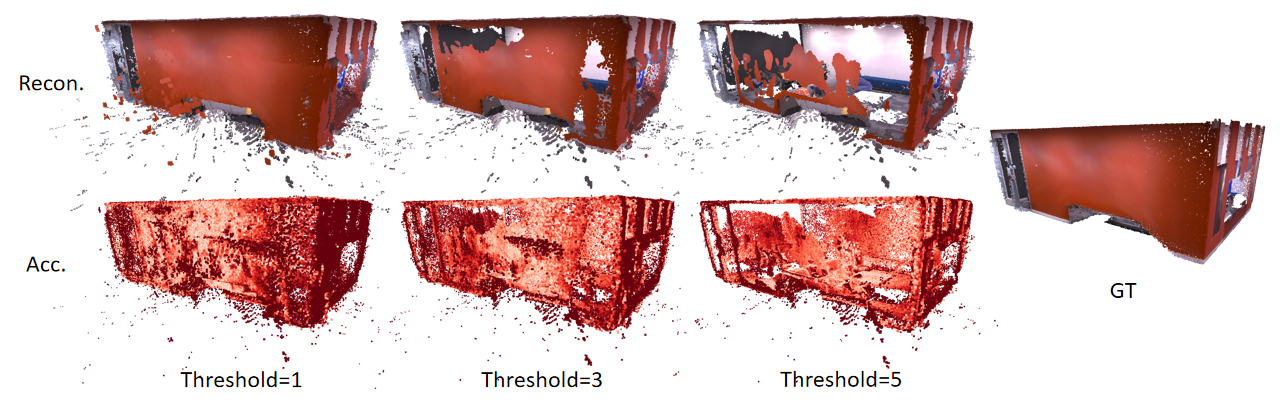}
\caption{
The reconstruction results and the corresponding accuracy heatmaps of MASt3R~\cite{leroy2024grounding} on Office 3 from Replica~\cite{replica19arxiv} dataset under different confidence thresholds. Lighter colors indicate higher accuracy. 
}
\label{fig:master_failure}
\end{figure*}

%% file: fig/incremental.tex
\begin{figure*}[!th]
\centering
\includegraphics[width=1.0\linewidth]{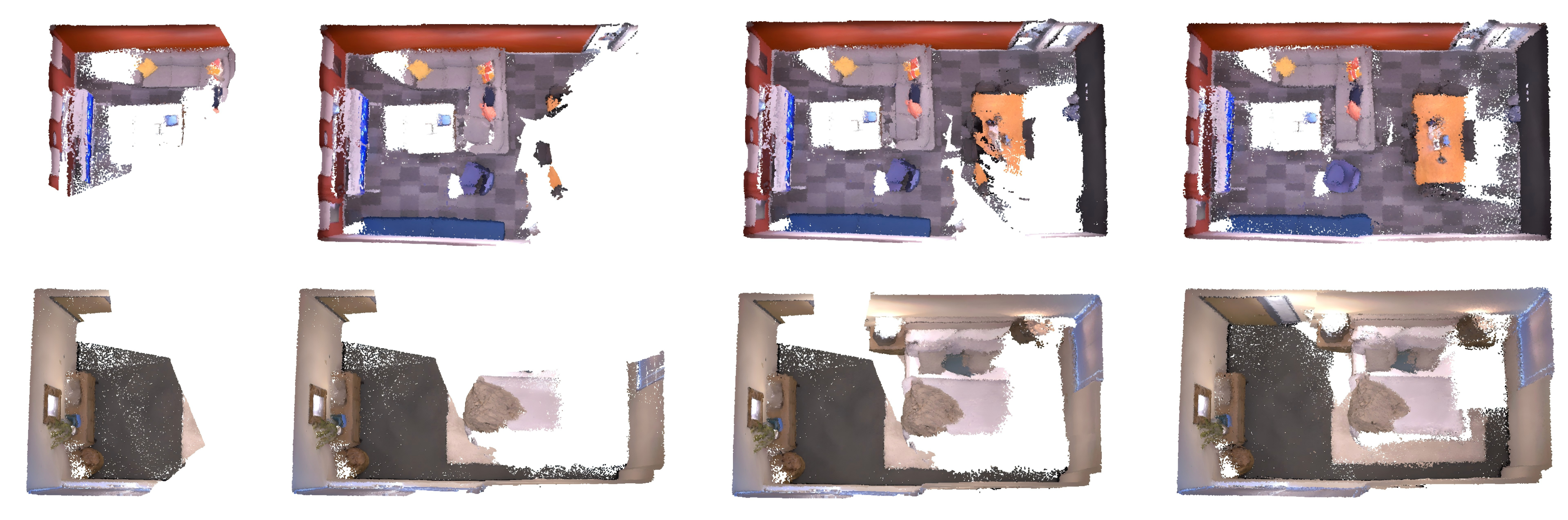}
\caption{
Visualization of the incremental reconstruction process of our method on the Office 3 and Room 1 of Replica~\cite{replica19arxiv} dataset. Our method achieves low drift without any global-optimization stage.
}
\label{fig:incremental}
\end{figure*}

%% file: fig/dtu.tex
\begin{figure*}[!th]
\centering
\includegraphics[width=0.98\linewidth]{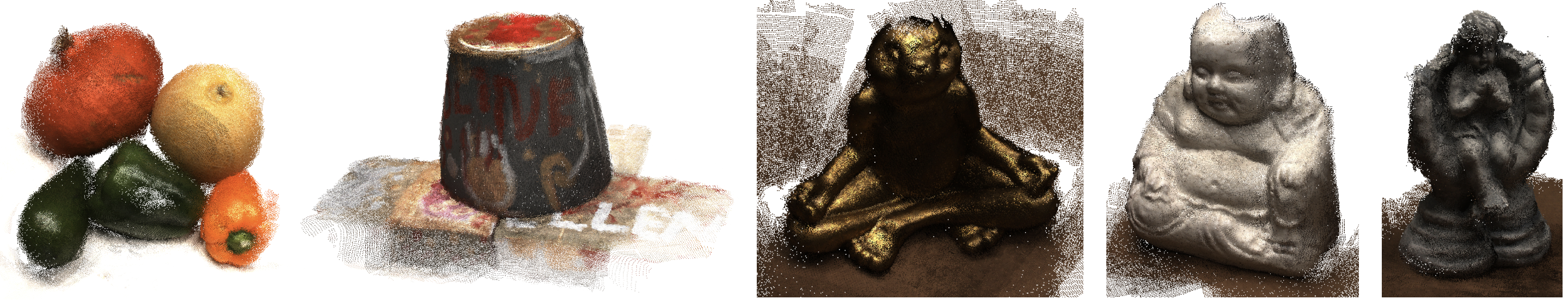}
\caption{
Reconstruction results on unorganized image collections from DTU~\cite{aanaes2016large} dataset.}
\label{fig:dtu}
\end{figure*}

%% file: tab/7s_cam.tex
\begin{table*}[th!]
\centering

\resizebox{0.87\textwidth}{!}{

\small

\begin{tabular}{l|ccccccc|c|c} 
\toprule
{Method} & Chess & Fire & Heads & Office & Pumpkin & RedKitchen & Stairs & Average & {FPS} \\ 

\toprule
 
DUSt3R~\cite{wang2024dust3r} (w/PnP) & 5.09 & 4.88 & 2.52 & 12.07 & \textbf{10.64} & 10.35 & 10.55 & 8.02 & \cellcolor{color7!25} $<$1 \\ 
 
MASt3R~\cite{leroy2024grounding} (w/PnP) & 4.32 & \textbf{2.92} & \textbf{1.47} & 12.37 & 11.82 & 7.98 & \textbf{3.04} & \textbf{6.28} & \cellcolor{color8!25} $\ll$1 \\ 

 NICER-SLAM~\cite{zhu2024nicer}* & \textbf{3.28} & 6.85 & 4.16 & \textbf{10.84} & 20.00 & \textbf{3.94} & 10.81 & 8.55 & \cellcolor{color7!25} $<$1 \\ 

\midrule

 DROID-SLAM~\cite{teed2021droid}* & \textbf{3.36} & \textbf{2.40} & \textbf{1.43} & \textbf{9.19} & 16.46 & \textbf{4.94} & \textbf{1.85} & \textbf{5.66} & \cellcolor{color3!25} $\sim$20 \\

 Spann3R~\cite{wang20243d} & 9.18 & 6.69 & 7.10 & 21.56 & 12.83 & 14.06 & 10.43 & 11.70 & \cellcolor{color1!25} $>$50 \\

 \textbf{SLAM3R-NoConf (Ours)} & 6.29 & 5.33 & 4.47 & 12.42 & 11.74 & 9.53 & 9.30 & 8.44 & \cellcolor{color2!25} $\sim$25 \\

 \textbf{SLAM3R (Ours)} & 6.20 & 5.30 & 4.56 & 12.40 & \textbf{11.71} & 9.47 & 9.20 & 8.41 & \cellcolor{color2!25} $\sim$25 \\

\bottomrule
\end{tabular}
}
\caption{
Camera pose estimation results on the 7Scenes~\cite{shotton2013scene} dataset reported using the ATE-RMSE (cm) metric. The average numbers are computed over all test scenes. * denotes the results reported in NICER-SLAM. 
}
\label{tab:7s_cam}
\end{table*}

%% file: tab/replica_cam.tex
\begin{table*}[th!]
\centering

\resizebox{0.96\textwidth}{!}{

\small

\begin{tabular}{l|cccccccc|c|c} 
\toprule

{Method}& Room 0 & Room 1 & Room 2 & Office 0 & Office 1 & Office 2 & Office 3 & Office 4 & Average & {FPS} \\

\toprule

DUSt3R~\cite{wang2024dust3r} (w/PnP) & 4.00 & 4.49 & 7.62 & 4.88 & 4.04 & 3.90 & 2.84 & 6.30 & 4.76 & \cellcolor{color7!25} $<$1 \\ 

MASt3R~\cite{leroy2024grounding} (w/PnP) & \textbf{1.07} & \textbf{0.99} & \textbf{0.87} & \textbf{0.90} & 4.90 & \textbf{1.21} & 1.77 & \textbf{1.63} & \textbf{1.67} & \cellcolor{color8!25} $\ll$1 \\ 

NICER-SLAM~\cite{zhu2024nicer}* & 1.36 & 1.60 & 1.14 & 2.12 & \textbf{3.23} & 2.12 & \textbf{1.42} & 2.01 & 1.88 & \cellcolor{color7!25} $<$1 \\ 

\midrule

GO-SLAM~\cite{zhang2023go} & - & - & - & - & - & - & - & - & 0.39 & \cellcolor{color4!25} $\sim$8 \\

DIM-SLAM~\cite{li2023dense} & 0.48 & 0.78 & 0.35 & 0.67 & \textbf{0.37} & 0.36 & \textbf{0.33} & \textbf{0.36} & 0.46 & \cellcolor{color5!25} $\sim$3 \\

DROID-SLAM~\cite{teed2021droid}* & \textbf{0.34} & \textbf{0.13} & \textbf{0.27} & \textbf{0.25} & 0.42 & \textbf{0.32} & 0.52 & 0.40 & \textbf{0.33} & \cellcolor{color3!25} $\sim$20 \\

Spann3R~\cite{wang20243d}  & 29.76 & 34.78 & 26.08 & 34.50 & 22.65 & 34.47 & 42.24 & 37.84 & 32.79 & \cellcolor{color1!25} $>$50 \\

\textbf{SLAM3R-NoConf (Ours)} & 4.54 & 5.89 & 5.73 & 11.17 & 6.32 & 6.15 & 4.99 & 8.05 & 6.61 & \cellcolor{color2!25} $\sim$24 \\

\textbf{SLAM3R (Ours)} & 4.56 & 5.88 & 5.72 & 11.17 & 6.32 & 6.15 & 4.95 & 8.09 & 6.61 & \cellcolor{color2!25} $\sim$24 \\

\bottomrule
\end{tabular}
}
\caption{
Camera pose estimation results on the Replica~\cite{replica19arxiv} dataset reported using the ATE-RMSE (cm) metric. 
}
\label{tab:replica_cam}
\end{table*}

%% file: tab/scannet.tex
\begin{table}[htb!]
\centering
\resizebox{0.48\textwidth}{!}{
\begin{tabular}{l|cccc}
\toprule
ScanNet & scene0011\_00 & scene0015\_00 & scene0019\_00 & Average   \\ 
\toprule
DUSt3R~\cite{wang2024dust3r}  & \textbf{5.56} / \textbf{3.76} & \textbf{5.04} / \textbf{4.10} & 4.52 / 4.74 & \textbf{5.04} / \textbf{4.20} \\  
Spann3R~\cite{wang20243d}  & 13.09 / 11.37 & 8.51 / 7.79 & 7.97 / 9.66 & 9.86 / 9.61 \\
\textbf{SLAM3R (Ours)} & 5.86 / 3.98 & 5.98 / 5.97 & \textbf{4.27} / \textbf{4.34} & 5.37 / 4.76 \\
\toprule
Tanks and Temples & Ignitius & Truck & Caterpillar & Average   \\ 
\toprule
DUSt3R~\cite{wang2024dust3r}  & 3.55/ 1.22 & 9.31 / \textbf{4.85} & 12.67 / 5.25 & 8.51 / \textbf{3.77} \\ 
Spann3R~\cite{wang20243d}  & 5.51 / 1.10 & 6.40 / 12.61 & \textbf{11.50}/ 5.74 & 7.80 / 6.48 \\
\textbf{SLAM3R (Ours)} & \textbf{3.30} / \textbf{0.94} & \textbf{5.35}/ 5.59 & 12.26 / \textbf{5.05} & \textbf{6.97} / 3.86 \\
\toprule
ETH3D & plant\_scene\_1 & table\_3 & sofa\_1 & Average   \\ 
\toprule
DUSt3R~\cite{wang2024dust3r}  & 2.98 / 2.48 & 3.13 / \textbf{1.30} & \textbf{2.05} / 3.67 & 2.72 / 2.48 \\ 
Spann3R~\cite{wang20243d}  & 2.54 / 4.25 & 3.03 / 2.08 & 2.10 / 4.55 & 2.56 / 3.62 \\
\textbf{SLAM3R (Ours)} & \textbf{2.36} / \textbf{1.98} & \textbf{2.75} / 1.34 & 2.13 / \textbf{1.90} & \textbf{2.41} / \textbf{1.74} \\
\bottomrule
\end{tabular}
}
\caption{
Reconstruction errors (accuracy / completeness) on ScanNet~\cite{dai2017scannet}, Tanks and Temples~\cite{knapitsch2017tanks}, and ETH3D~\cite{schops2019bad} datasets. }
\label{tab:reb_scannet}
\end{table}

%% file: fig/fig_diminishing_return.tex
\begin{figure}[!t]\centering
    \includegraphics[width=9cm]    {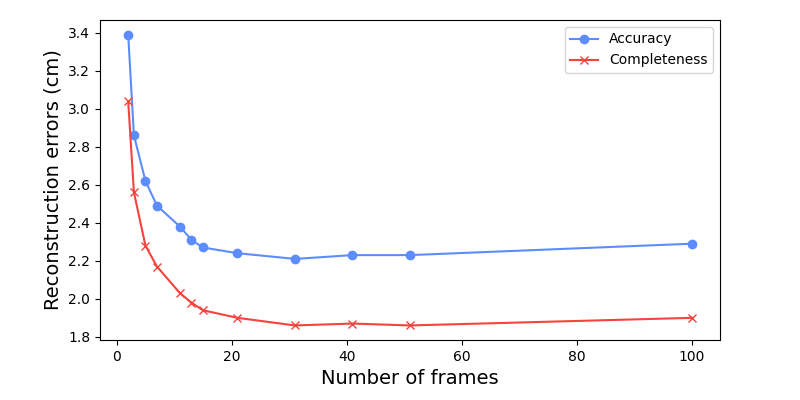}
	\caption{Inner-window keyframe reconstruction results from various window lengths.}
    \label{fig:diminishing_return}
\end{figure}

%% file: tab/sceneframe_num.tex
\begin{table}
    \centering
    \footnotesize

    \begin{tabular}{lcccc}
    \toprule
         \# Scene frames & Acc. & Comp. & FPS  \\
    \toprule
         1 & 4.18 & 2.61 & \cellcolor{color1!25} $\sim$398 \\
         5 & 3.99 & 2.79 & \cellcolor{color2!25} $\sim$247 \\
         10 & \textbf{3.57} & 2.62 & \cellcolor{color3!25} $\sim$152 \\
         20 & \textbf{3.57} & 2.60 & \cellcolor{color4!25} $\sim$86 \\
         30 & 3.59 & \textbf{2.58} & \cellcolor{color5!25} $\sim$61 \\
         40 & 4.15 & 3.05 & \cellcolor{color6!25} $\sim$46 \\
         50 & 4.27 & 3.15 & \cellcolor{color7!25} $\sim$37 \\
    \bottomrule
    \end{tabular}
\caption{
Reconstruction results on Replica~\cite{replica19arxiv} dataset, with various maximum number of scene frames selected for keyframe registration. The FPS of the L2W model aligning 10 keyframes at once with different numbers of input scene frames is also reported.
}
\label{tab:scene_frame_num}
\end{table}